\documentclass[runningheads]{llncs}
\usepackage{graphicx}
\usepackage{amssymb,amsmath,graphicx,float,adjustbox,array}
\usepackage{color}
\usepackage{graphicx}
\usepackage{xfrac}
\usepackage{subfig}
\usepackage{enumitem}
\usepackage{hyperref}
\usepackage{breakcites}

\begin{document}
\pagestyle{headings}
\mainmatter
\newcommand{\point}{
    \raise0.7ex\hbox{.}
    }

\title{MoDeep: A Deep Learning Framework Using Motion Features for Human Pose Estimation}

\titlerunning{MoDeep: A Deep Learning Framework Using Motion Features}

\authorrunning{Arjun Jain, Jonathan Tompson, Yann LeCun and Christoph Bregler\\
}

\author{
Arjun Jain, Jonathan Tompson, Yann LeCun and Christoph Bregler\\
}

\institute{
New York University \\
\texttt{\{ajain, tompson, yann, bregler\}@cs.nyu.edu} \\
}

\maketitle

\begin{abstract}
In this work, we propose a novel and efficient method for articulated human pose estimation in videos using a convolutional network architecture, which incorporates both color and motion features. We propose a new human body pose dataset, \emph{FLIC-motion}\footnote{This dataset can be downloaded from \url{http://cs.nyu.edu/~ajain/accv2014/}}, that extends the FLIC dataset~\cite{modec} with additional motion features. We apply our architecture to this dataset and report significantly better performance than current state-of-the-art pose detection systems.
\end{abstract}

\section{Introduction}

Human body pose recognition in video is a long-standing problem in computer vision with a wide range of applications. However, body pose recognition remains a challenging problem due to the high dimensionality of the input data and the high variability of possible body poses. Traditionally, computer vision-based approaches tend to rely on appearance cues such as texture patches, edges, color histograms, foreground silhouettes or hand-crafted local features (such as histogram of gradients (HoG)~\cite{Dalal}) rather than motion-based features. Alternatively, psychophysical experiments~\cite{biologicalmotion} have shown that motion is a powerful visual cue that alone can be used to extract high-level information, including articulated pose. 

Previous work~\cite{Ferrari08,weiss:sidestepping} has reported that using motion features to aid pose inference has had little or no impact on performance. Simply adding high-order temporal connectivity to traditional models would most often lead to intractable inference.  In this work we show that deep learning is able to successfully incorporate motion features and is able to out-perform existing state-of-the-art techniques. Further, we show that by using motion features alone our method outperforms~\cite{Eichner:2009:BAM,yang11cvpr,sapp11eccv} (see Fig~\ref{fig:flic_results}(a) and (b)), which further strengthens our claim that information coded in motion features is valuable and should be used when available.

This paper makes the following contributions:
\begin{itemize}[topsep=\parskip]
\item A system that successfully incorporates motion-features to enhance the performance of pose-detection `in-the-wild' compared to existing techniques.
\item An efficient and tractable algorithm that achieves close to real-time frame rates, making our method suitable for wide variety of applications.
\item A new dataset called {\bf FLIC-motion}, which is the FLIC dataset~\cite{modec} augmented with `motion-features' for each of the 5003 images collected from Hollywood movies. 
\end{itemize}

\section{Prior Work}

\paragraph{\bf Geometric Model Based Tracking:}
One of the earliest works on articulated tracking in video was Hogg~\cite{hogg1983model} in 1983 using edge features and a simple cylinder based body model.  Several other model based articulated tracking systems have been reported over the past two decades, most notably~\cite{rehg1995model,kakadiaris1996model,wren1997pfinder,bregler1998tracking,deutscher2000articulated,sidenbladh2000stochastic,sminchisescu2001covariance}.  The models used in these systems were explicit 2D or 3D jointed geometric models.  Most systems had to be hand-initialized (except~ \cite{wren1997pfinder}), and focused on incrementally updating pose parameters from one frame to the next. More complex examples come from the HumanEva dataset competitions~\cite{Sigal2010} that use video or higher-resolution shape models such as SCAPE~ \cite{anguelov2005scape} and extensions. We refer the reader to~\cite{poppe2007vision} for a complete survey of this era.   Most recently such techniques have been shown to create very high-resolution animations of detailed body and cloth deformations \cite{de2008performance,jain2010moviereshape,stoll2011fast}.  Our approach differs, since we are dealing with single view videos in unconstrained environments.

\paragraph{\bf Statistical Based Recognition:}
One of the earliest systems that used no explicit geometric model was reported by Freeman et al. in 1995 \cite{freeman1995orientation} using oriented angle histograms to recognize hand configurations.  This was the precursor for the bag-of-features, SIFT~\cite{SIFT}, STIP~\cite{STIP}, HoG, and Histogram of Flow (HoF)~\cite{HOF} approaches that boomed a decade later, most notably including the work by Dalal and Triggs in 2005 \cite{dalal2005histograms}. Different architectures have since been proposed, including ``shape-context'' edge-based histograms from the human body \cite{mori2002estimating,agarwal2006recovering} or just silhouette features \cite{Grauman2003}.  Shakhnarovich et al. \cite{Shakhnarovich2003} learn a parameter sensitive hash function to perform example-based pose estimation. Many techniques have been proposed that extract, learn, or reason over entire body features, using a combination of local detectors and structural reasoning (see \cite{ramanan2005strike} for coarse tracking and \cite{buehler2009learning} for person-dependent tracking).

Though the idea of using ``Pictorial Structures"  by Fischler and Elschlager~\cite{Fischler73} has been around since the 1970s, matching them efficiently to images has only been possible since the famous work on `Deformable Part Models' (DPM) by Felzenszwalb et al.~\cite{felzenszwalb2008discriminatively} in 2008. Many algorithms that use DPM for creating the body part unary distribution~\cite{andriluka2009pictorial,Eichner:2009:BAM,yang11cvpr,dantone13cvpr} with spatial-models incorporating body-part relationship priors have since then been developed. 
Johnson and Everingham~\cite{johnson11cvpr}, who also proposed the `Leeds Sports Database', employ a cascade of body part detectors to obtain more discriminative templates. Almost all best performing algorithms since have solely built on HoG and DPM for local evidence, and yet more sophisticated spatial models.  Pishchulin~\cite{pishchulin13cvpr} proposes a model that augments the DPM unaries with \emph{Poselet} conditioned \cite{PoseletsICCV09} priors. Sapp and Taskar~\cite{modec} propose a model where they cluster images in the pose-space and then find the mode which best describes the input image. The pose of this mode then acts as a strong spatial prior, whereas the local evidence is again based on HoG and gradient features. Following the \emph{Poselets} approach \cite{PoseletsICCV09}, the \emph{Armlets} approach by Gkioxari et al.~\cite{Gkioxari:2013:APE} incorporates edges, contours, and color histograms in addition to the HoG features. They employ a semi-global classifier for part configuration and show good performance on real-world data. However, they only show their results on arms. The major drawback of all these approaches is that both the local evidence and the global structure is hand crafted, whereas we jointly learn both the local features and the global structure using a multi-resolution convolutional network.

Shotton et al.~\cite{shotton2013real} use an ensemble of random trees to perform per-pixel labeling of body parts in depth images.  As a means of reducing overall system latency and avoiding repeated false detections, their work focuses on pose inference using only a single depth image.  By contrast, we extend the single frame requirement to at least 2 frames (which we show considerably improves pose inference), and our input domain is unconstrained RGB images rather than depth.

\paragraph{\bf Pose Detection Using Image Sequences:}

\paragraph{\bf Deep Learning based Techniques:}

Recently, state-of-the-art performance has been reported on many vision tasks using deep learning algorithms~\cite{ZeilerSuperVision04,razavian2014cnn,DeepFaceCVPR2014,deng2013deep,pedestrianCVPR13,deepflow}. \cite{deeppose,jainiclr2014,tompsonTOG14} also apply neural networks for pose recognition, specifically Toshev et al.~\cite{deeppose} show better than state-of-the-art performance on the `FLIC' and `LSP'~\cite{Johnson10} datasets. In contrast to Toshev et al., in our work we propose a translation invariant model which improves upon their method, especially in the high-precision region.

\section{Body-Part Detection Model} 

We propose a Convolutional Network (ConvNet) architecture for the task of estimating the 2D location of human joints in video (section~\ref{sec:conv_model}). The input to the network is an RGB image and a set of  \emph{motion features}.  We investigate  a wide variety of motion feature formulations (section~\ref{sec:motionFeats}).  Finally, we will also introduce a simple Spatial-Model to solve a specific sub-problem associated with evaluation of our model on the FLIC-motion dataset (section~\ref{sec:spatialModel}).

\subsection{Motion Features}
\label{sec:motionFeats}


The aim of this section is to incorporate features that are representative of the true \emph{motion-field} (the perspective projection of the 3D velocity-field of moving surfaces) as input to our detection network so that it can exploit motion as a cue for body part localization. To this end, we evaluate and analyze four motion features which fall under two broad categories: those using simple derivatives of the RGB video frames and those using optical flow features.  For each RGB image pair $f_{i}$ and $f_{i+\delta}$, we propose the following features:

\begin{itemize}
\item RGB image pair - $\left\{f_{i}, f_{i+\delta}\right\}$
\item RGB image and an RGB difference image - $\left\{f_{i}, f_{i+\delta} - f_{i}\right\}$
\item Optical-flow\footnote{We use the algorithm proposed by Weinzaepfel et al.~\cite{deepflow} to compute optical-flow.} vectors - $\left\{f_{i},\operatorname{FLOW}(f_{i}, f_{i+\delta})\right\}$
\item Optical-flow magnitude - $\left\{f_{i},||\operatorname{FLOW}(f_{i}, f_{i+\delta})||_2\right\}$
\end{itemize}

The RGB image pair is by far the simplest way of incorporating the relative motion information between the two frames.  However, this representation clearly suffers from a lot of redundancy (i.e. if there is no camera movement) and is extremely high dimensional.  Furthermore, it is not obvious to the deep network what changes in this high dimensional input space are relevant temporal information and what changes are due to noise or camera motion. A simple modification to this representation is to use a difference image, which reformulates the RGB input so that the algorithm sees directly the pixel locations where high energy corresponds to motion (alternatively the network would have to do this implicitly on the image pair).  A more sophisticated representation is optical-flow, which is considered to be a high-quality approximation of the true \emph{motion-field}. Implicitly learning to infer optical-flow from the raw RGB input would be non-trivial for the network to estimate, so we perform optical-flow calculation as a pre-processing step (at the cost of greater computational complexity).

\subsubsection{FLIC-motion dataset:}
\label{sec:FLICmotion}
We propose a new dataset which we call {\bf FLIC-motion}\footnote{This dataset can be downloaded from \url{http://cs.nyu.edu/~ajain/accv2014/}}. It is comprised of the original FLIC dataset of 5003 labeled RGB images collected from 30 Hollywood movies, of which 1016 images are held out as a test set, augmented with the aforementioned motion features.

\begin{figure}[th]
  \centering
  \subfloat[\label{fig:nonw_avg}]{
        \includegraphics[height=.13\textheight]{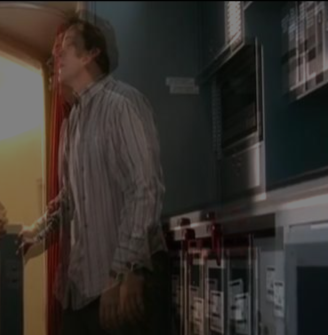}
  }    
  \subfloat[\label{fig:nonw_flow}]{
        \includegraphics[height=.13\textheight]{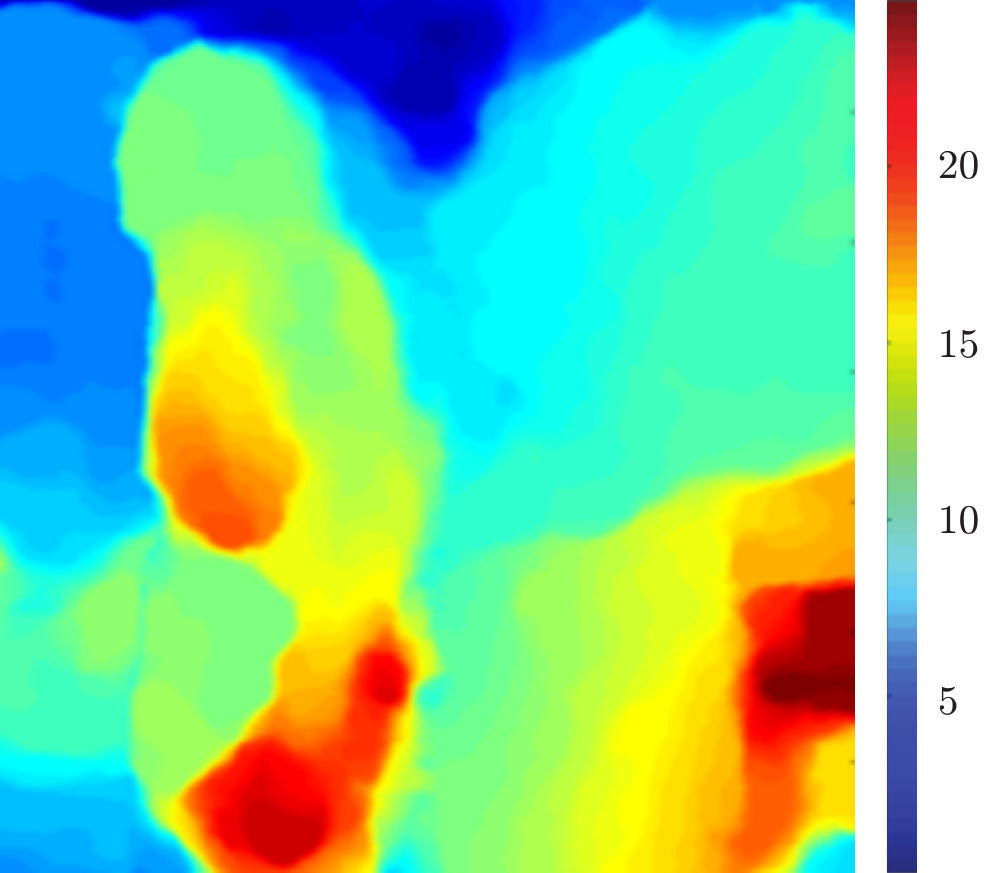}
  }  
  \subfloat[\label{fig:w_avg}]{
        \includegraphics[height=.13\textheight]{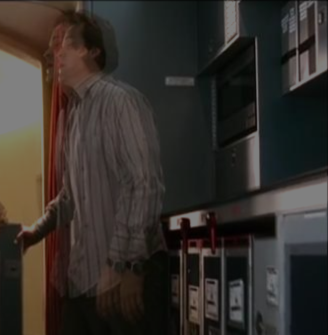}
  }
  \subfloat[\label{fig:w_flow}]{
        \includegraphics[height=.13\textheight]{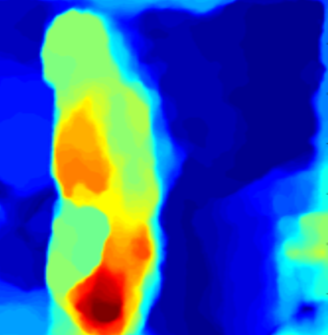}
  }

  \caption{Results of optical-flow computation: (a) average of frame pair, (b) optical flow on (a), (c) average of frame pair after camera compensation, and (d) optical-flow on (c)}
  \label{fig:warpdemo}
\end{figure}

We experimented with different values for $\delta$ and investigated the above features with and without camera motion compensation; we use a simple 2D projective motion model between $f_{i}$ and $f_{i+\delta}$, and warp $f_{i+\delta}$ onto $f_i$ using the inverse of this best fitting projection to approximately remove camera motion. A comparison between image pairs with and without warping can be seen in~Fig~\ref{fig:warpdemo}.

To obtain $f_{i+\delta}$, we must know where the frames $f_{i}$ occur in each movie. Unfortunately, this was non-trivial as the authors Sapp et al.~\cite{modec} could not provide us with the exact version of the movie that was used for creating the original dataset. Corresponding frames can be very different in multiple versions of the same movie (4:3 vs wide-screen, director's cut, special editions, etc.). We estimate the best similarity transform $S$ between $f_{i}$ and each frame $f^{m}_{j}$ from the movie $m$, and if the distance $|f_{i} - S f^{m}_{j}|$ is below a certain threshold (10 pixels), we conclude that we found the correct frame. We visually confirm the resulting matches and manually pick frames for which the automatic matching was unsuccessful (e.g. when enough feature points were not found).

\subsection{Convolutional Network}
\label{sec:conv_model}

Recent work~\cite{deeppose,jainiclr2014} has shown ConvNet architectures are well suited for the task of human body pose detection, and due to the availability of modern Graphics Processing Units (GPUs), we can perform Forward Propagation (FPROP) of deep ConvNet architectures at interactive frame-rates. Similarly, we realize our detection model as a deep ConvNet architecture. The input is a 3D tensor containing an RGB image and its corresponding motion features, and the output is a 3D tensor containing \emph{response-maps}, with one response-map for each joint. Each response-map describes the per-pixel energy for the presence of the corresponding joint at that pixel location.  

\begin{figure}[th]
\centering
    \includegraphics[width=\columnwidth]{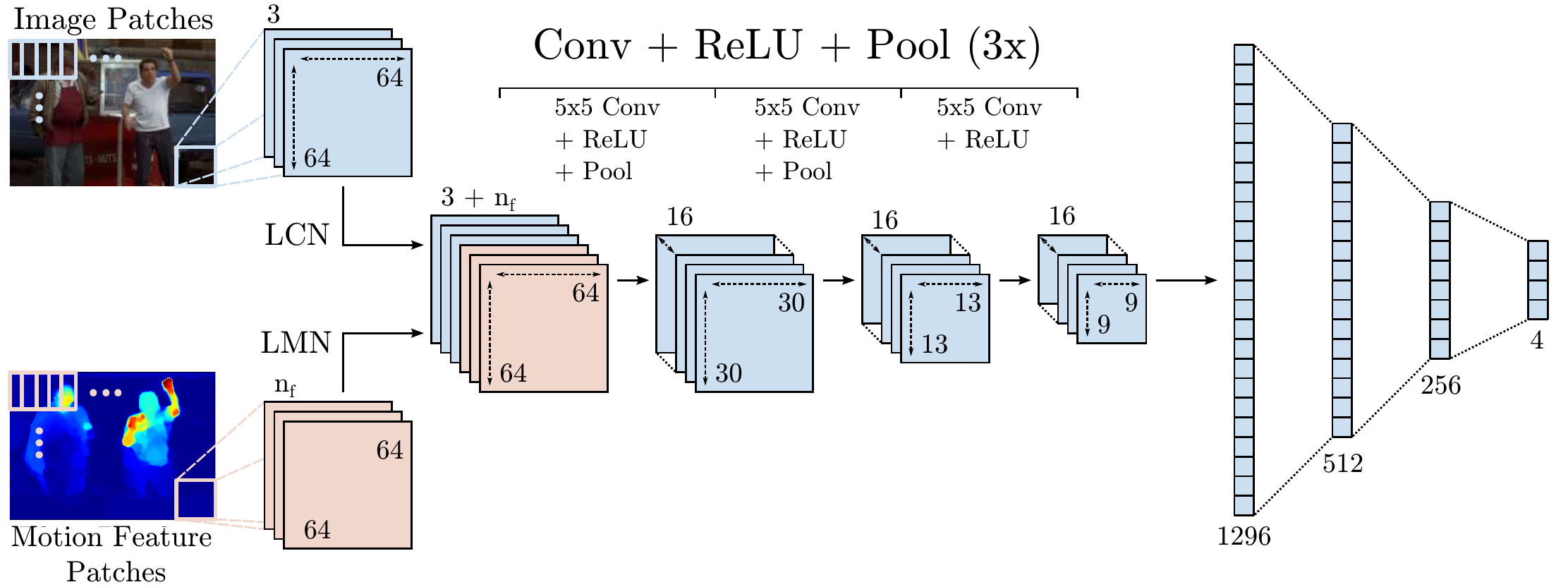}
    \caption{Sliding-window with image and flow patches}
  \label{fig:multiResPatchModel}
\end{figure}

Our ConvNet is based on a \emph{sliding-window} architecture.  A simplified version of this architecture is shown in Fig~\ref{fig:multiResPatchModel}. The input patches are first normalized using Local Contrast Normalization (LCN~\cite{torch7}) for the RGB channels and a new normalization method for the motion features we call \emph{Local Motion Normalization} (LMN). We formulate LMN as the local subtraction with the response from a Gaussian kernel with large standard deviation followed by a divisive normalization.  The result is that it removes some unwanted background camera motion as well as normalizing the local intensity of motion (which helps improve network generalization for motions of varying velocity but with similar pose).  
Prior to processing through the convolution stages, the normalized motion channels are concatenated along the feature dimension with the normalized RGB channels, and the resulting tensor is processed though 3 stages of convolution. 

The first two convolution stages use rectified linear units (ReLU) and Max-pooling, and the last stage incorporates a single ReLU layer. The output of the last convolution stage is then passed to a three stage fully-connected neural-network. The network is then applied to all $64\times64$ sub-windows of the image, stepped every 4 pixels horizontally and vertically. This produces a dense response-map output, one for each joint. The major advantage of this model is that the learned detector is translation invariant by construction.

\begin{figure}[th]
  \centering
    \includegraphics[width=\columnwidth]{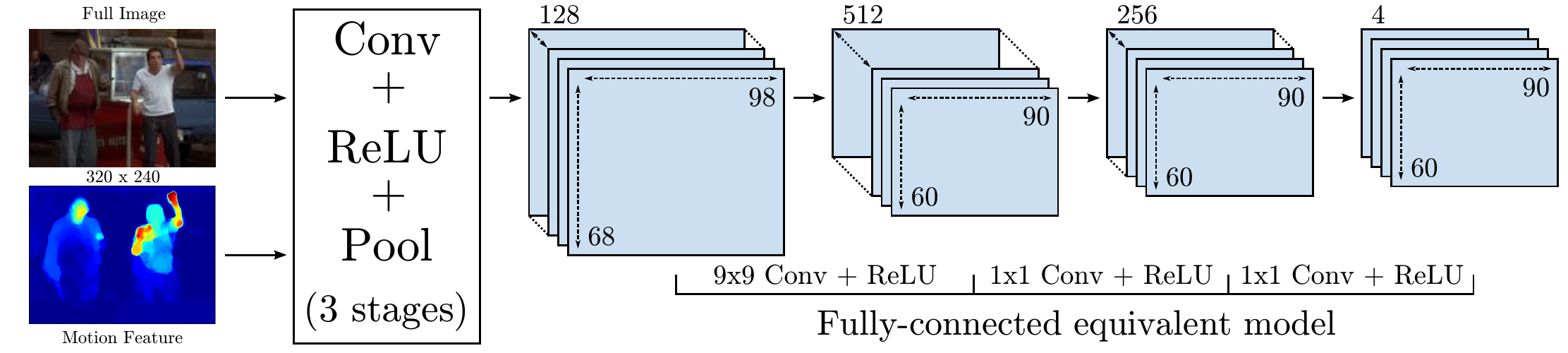}
    \caption{Efficient sliding window model}
  \label{fig:oneShotOneBank}
\end{figure}

Because the layers are convolutional, applying two instances of the network in Fig~\ref{fig:multiResPatchModel} to two overlapping input windows leads to a considerable amount of redundant computation. Recent work~\cite{fastcnn,overfeatSermanet} eliminates this redundancy and thus yields a dramatic speed up. This is achieved  by applying each layer of the convolutional network to the entire input image. The fully connected layers for each window are also replicated for all sub-windows of the input.  This formulation allows us to back-propagate though this network for all windows simultaneously. Due to the two $2\times2$ subsampling layers, we obtain one output vector every $4\times4$ input pixels. An equivalent efficient version of the sliding window model is shown in Fig~\ref{fig:oneShotOneBank}.  

Note that an alternative model (such as in Tompson et al.~\cite{tompsonTOG14}) would replace the last 3 convolutional layers with a fully-connected neural network whose input context is the feature activations for the entire input image.  Such a model would be appropriate if we knew a priori that there existed a strong correlation between skeletal pose and the position of the person in the input frame since this alternative model is not invariant with respect to the translation of the person within the image.  However, the FLIC dataset has no such strong pose-location bias (i.e. a subject's torso is not always in the same location in the image), and therefore a sliding-window based architecture is more appropriate for our task.

\begin{figure}[th]
  \centering
    \includegraphics[width=\columnwidth]{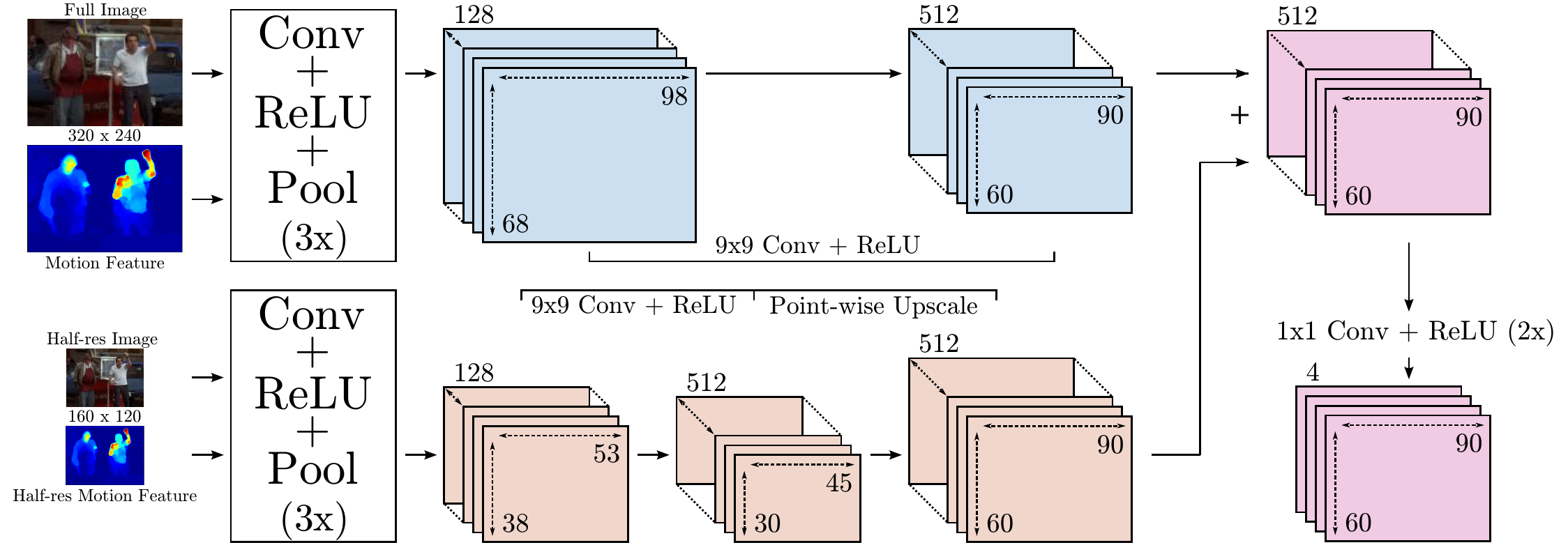}
    \caption{Multi-resolution efficient sliding window model}
  \label{fig:oneShotMultiBankSimplified}
\end{figure}

We extend the single resolution ConvNet architecture of Fig~\ref{fig:oneShotOneBank} by incorporating a \emph{multi-resolution} input. We do so by down-sampling the input (using appropriate anti-aliasing), and then each resolution image is processed through either a LCN or LMN layer using the same normalization kernels for each bank producing an approximate Laplacian pyramid.  The role of the Laplacian Pyramid is to provide each bank with non-overlapping spectral content which minimizes network redundancy.  Our final, multi-resolution network is shown in Fig~\ref{fig:oneShotMultiBankSimplified}.  The outputs of the convolution banks are concatenated (along the feature dimension) by point-wise up-scaling of the lower resolution bank to bring the feature maps into canonical resolution.  Note that in our final implementation we use 3 resolution banks.

We train the Part-Detector network using supervised learning via Back Propagation and Stochastic Gradient Descent.  We minimize a mean squared error criterion for the distance between the inferred response-map activation and a ground truth response-map, which is a 2D Gaussian distribution centered at the target joint location and with small standard deviation (1px).  We use Nesterov momentum to reduce training time~\cite{sutskeverimportance} and we randomly perturb the input images each epoch by randomly flipping and scaling the images to prevent network overtraining and improve generalization performance.

\subsection{Simple Spatial Model}
\label{sec:spatialModel}
Our model is evaluated on our new FLIC-motion dataset (section~\ref{sec:motionFeats}). As per the original FLIC dataset, the test images in FLIC-motion may contain multiple people, however, only a single actor per frame is labeled in the test set.  As such, a rough torso location of the labeled person is provided at test time to help locate the ``correct'' person.  We incorporate this information by means of a simple and efficient Spatial-Model.

\begin{figure}[th]
  \centering
    \includegraphics[width=\columnwidth]{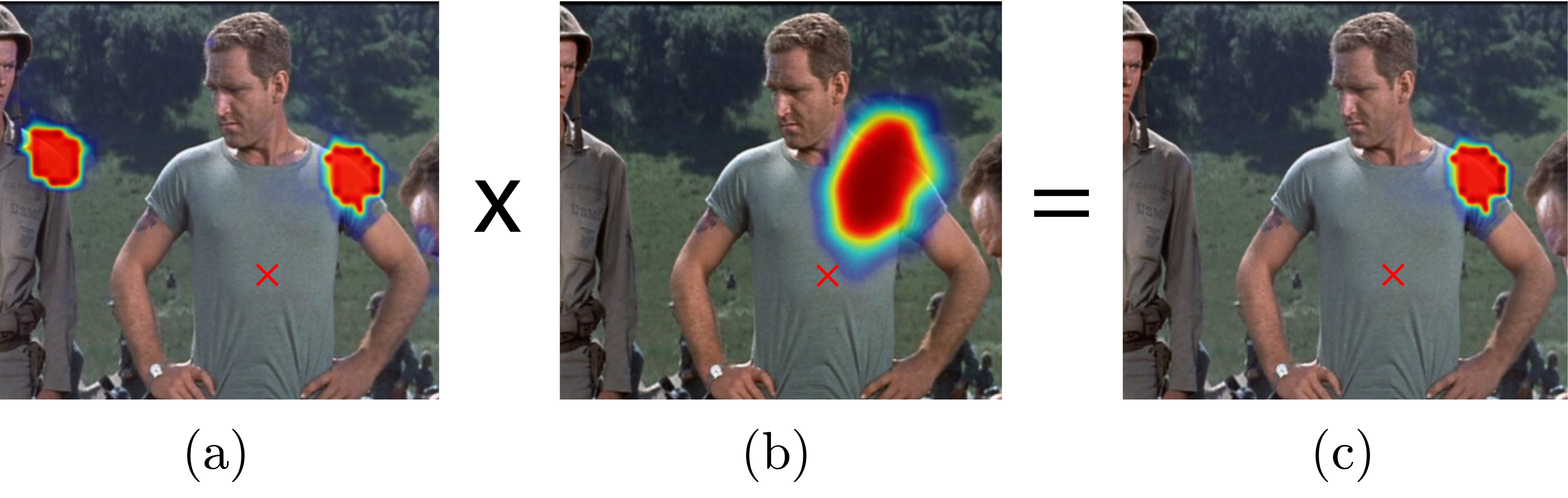}
    \caption{Simple spatial model used to mask-out the incorrect shoulder activations given a 2D torso position}
  \label{fig:spatialModel}
\end{figure}

The inclusion of this stage has two major advantages.  Firstly, the correct feature activation from the Part-Detector output is selected for the person for whom a ground-truth label was annotated.  An example of this is shown in Fig~\ref{fig:spatialModel}. Secondly, since the joint locations of each part are constrained in proximity to the single ground-truth torso location, then (indirectly) the connectivity between joints is also constrained, enforcing that inferred poses are anatomically viable (i.e. the elbow joint and the shoulder joint cannot be to far away from the torso, which in turn enforces spatial locality between the elbow and shoulder joints).

The core of our Spatial-Model is an empirically calculated \emph{joint-mask}, shown in Fig~\ref{fig:spatialModel}(b). The joint-mask layer describes the possible joint locations, given that the supplied torso position is in the center of the mask.  To create a mask layer for body part $A$, we first calculate the empirical histogram of the part $A$ location, $x_A$, relative to the torso position $x_T$ for the training set examples; i.e. $x_{\text{hist}}=x_A-x_T$.  We then turn this histogram into a Boolean mask by setting the mask amplitude to 1 for pixels for which $p\left(x_{\text{hist}}\right) > 0$. Finally, we blur the mask using a wide Gaussian low-pass filter which accounts for body part locations not represented in the training set (but which might be present in the test set).

During test time, this joint-mask is shifted to the ground-truth torso location and the per-pixel energy from the Part-Model (section~\ref{sec:conv_model}) is then multiplied with the mask to produce a filtered output.  This process is carried out for each body part independently.

It should be noted that while this Spatial-Model does enforce some anatomic consistency, it does have limitations.  Notably, we expect it to fail for datasets where the range of poses is not as constrained as the FLIC dataset (which is primarily front facing and standing up poses).


\section{Results}

Training time for our model on the FLIC-motion dataset (3957 training set images, 1016 test set images) is approximately 12 hours, and FPROP of a single image takes approximately 50ms\footnote{Analysis of our system was on a 12 core workstation with an NVIDIA Titan GPU}. For our models that use optical flow as a motion feature input,  the most expensive part of our pipeline is the optical flow calculation, which takes approximately 1.89s per image pair. (We plan to investigate real-time flow estimations in the future). 

Section~\ref{sec:resultsMotionFeats} compares the performance of the motion features from section~\ref{sec:motionFeats}.  Section~\ref{sec:resultsComparison} compares our architecture with other techniques and shows that our system significantly outperforms existing state-of-the-art techniques. Note that for all experiments in Section~\ref{sec:resultsMotionFeats} we use a smaller model with 16 convolutional features in the first 3 layers. A model with 128 instead of 16 features for the first 3 convolutional layers is used for results in Section~\ref{sec:resultsComparison}. 

\subsection{Comparison and Analysis of Proposed Motion Features}
\label{sec:resultsMotionFeats}

\begin{figure}[h]
  \centering
\includegraphics[width=\textwidth]{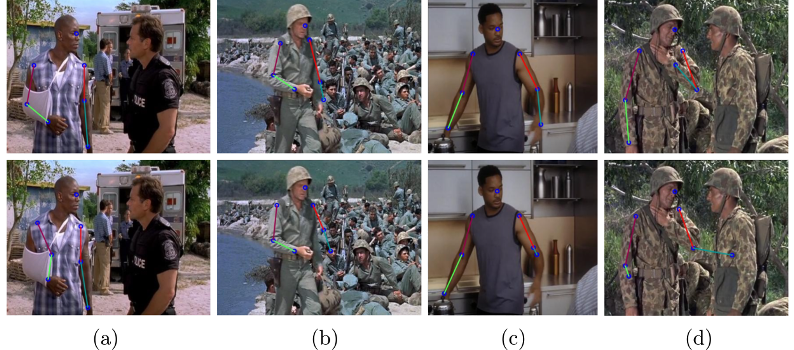}
 \caption{Predicted joint positions on the FLIC test-set. Top row: detection with motion features (L2 motion flow). Bottom row: without motion features (baseline).}
  \label{fig:with_without_motion}
\end{figure}

Fig~\ref{fig:with_without_motion} shows a selection of example images from the FLIC test set which highlights the importance of using motion features for body pose detection.  In Fig~\ref{fig:with_without_motion}(a), the elbow position is occluded by the actor's sling, and no such examples exist in the training set; however, the presence of body motion provides a strong cue for elbow location.  Figs~\ref{fig:with_without_motion}(b) and (d) have extremely cluttered backgrounds and the correct joint location is locally similar to the surrounding region (especially for the camouflaged clothing in Fig~\ref{fig:with_without_motion}(d)).  For these images, motion features are essential in correct joint localization.  Finally, Fig~\ref{fig:with_without_motion}(c) is an example where motion blur (due to fast joint motion) reduces the fidelity of RGB edge features, which results in incorrect localization when motion features are not used.

\begin{figure}[h]
  \centering
  \subfloat[FLIC-motion: Elbow\label{fig:features_flic_elbow}]{
        \includegraphics[width=.48\textwidth]{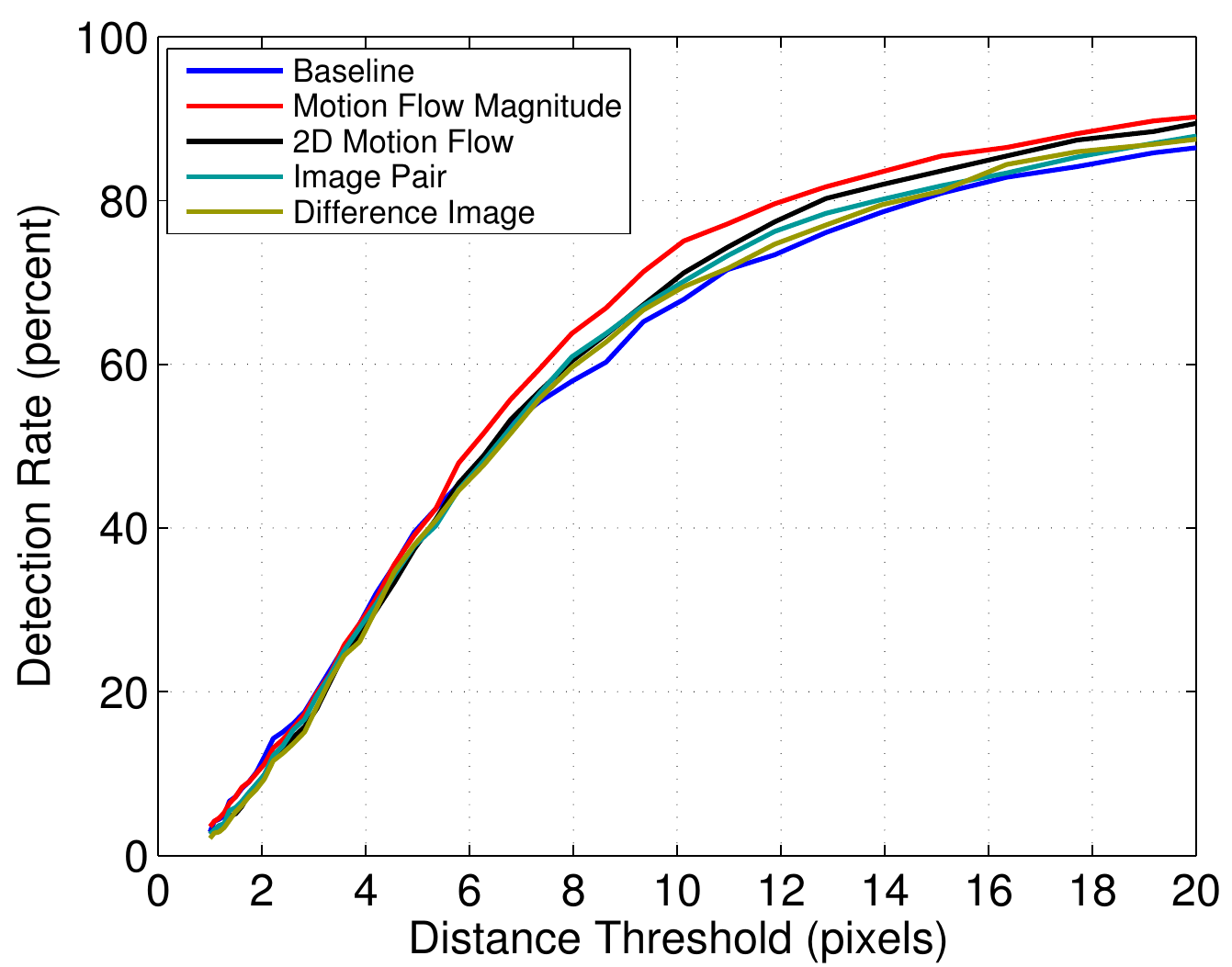}
   }
   \subfloat[FLIC-motion: Wrist\label{fig:features_flic_wrist}]{
        \includegraphics[width=.48\textwidth]{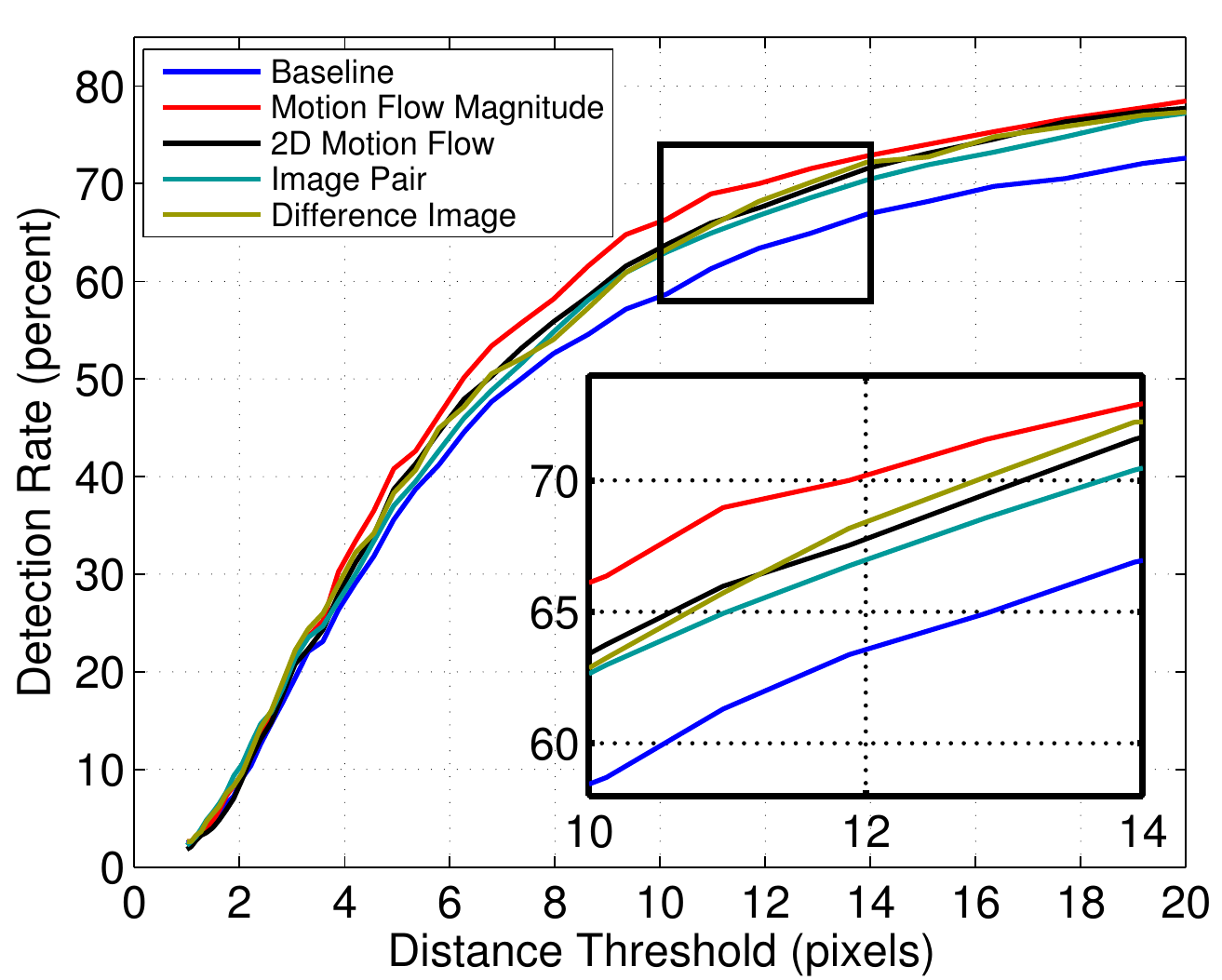}
   }
  \caption{Model performance for various motion features}
  \label{fig:features_flic_results}
\end{figure}

Figs~\ref{fig:features_flic_results}(a) and (b) show the performance of the motion features of section~\ref{sec:motionFeats} on the FLIC-motion dataset for the Elbow and Wrist joints respectively. For evaluating our test-set performance, we use the criterion proposed by Sapp et al.~\cite{modec}. We count the percentage of the test-set images where joint predictions are within a given radius that is normalized to a 100 pixel torso size. Surprisingly, even the simple frame-difference temporal feature improves upon the baseline result (which we define as a single RGB frame input) and even outperforms the 2D optical flow input (see \ref{fig:with_without_motion}(b) inset).

Note that stable and accurate calculation of optical-flow from arbitrary RGB videos is a very challenging problem.  Therefore, incorporating motion flow features as input to the network adds non-trivial localization cues that would be very difficult for the network to learn internally with limited learning capacity. Therefore, it is expected that the best performing networks in Fig~\ref{fig:features_flic_results} are those that incorporate motion flow features.  However, it is surprising that using the magnitude of the flow vectors performs as well as - and in some cases outperforms - the full 2D motion flow. Even though the input data is richer, we hypothesize that when using 2D flow vectors the network must learn invariance to the direction of joint movement; for instance, the network should predict the same head position  whether a person is turning his/her head to the left or right on the next frame.  On the other hand, when the L2 magnitude of the flow vector is used, the network sees the high velocity motion cue but cannot over-train to the direction of the movement.

\begin{figure}[h]
  \centering
  \subfloat[FLIC-motion: Wrist\label{fig:analysis_deltat}]{
        \includegraphics[width=.48\textwidth]{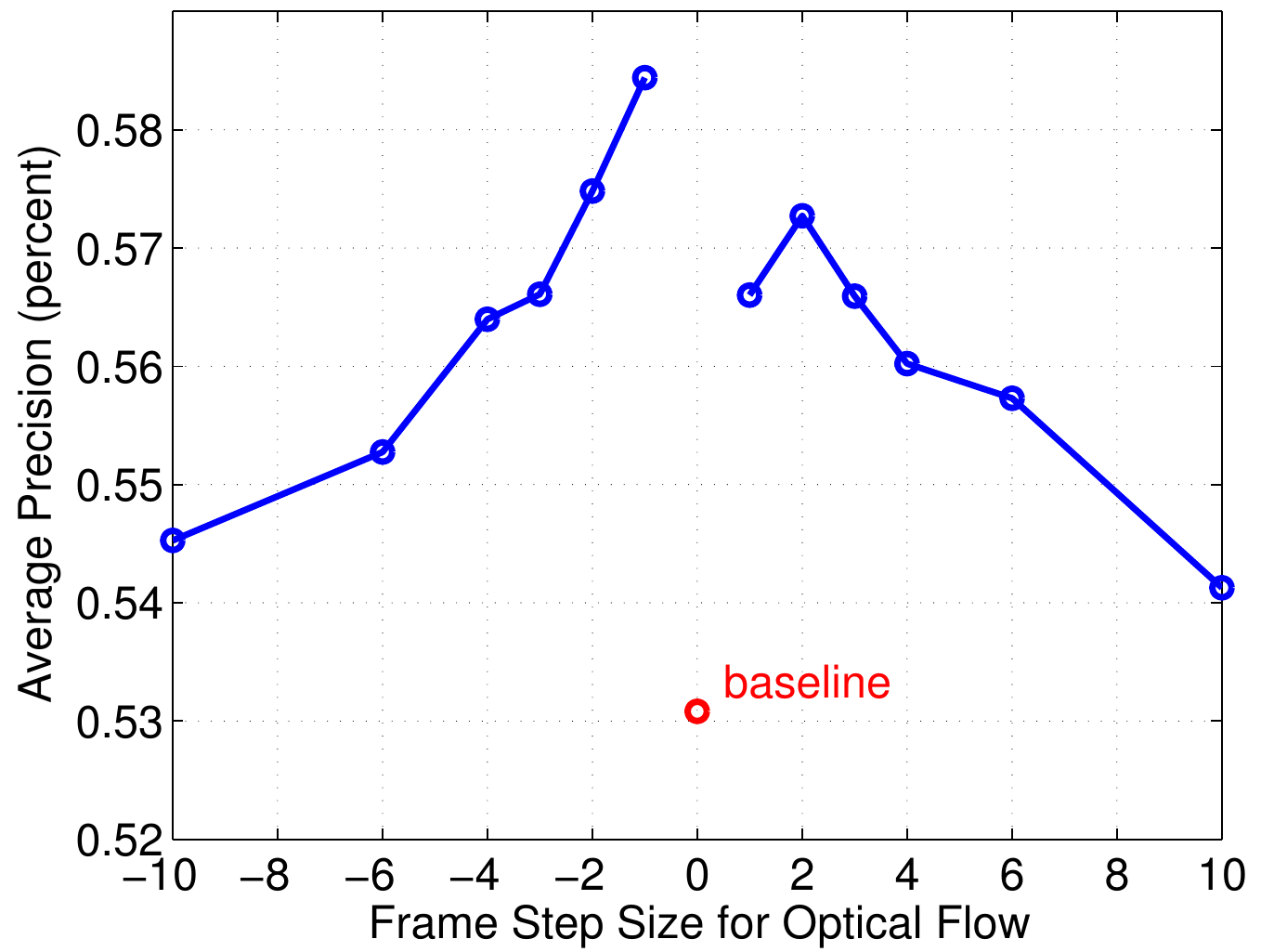}
  }    
  \subfloat[FLIC-motion: Wrist\label{fig:analysis_camera}]{
        \includegraphics[width=.48\textwidth]{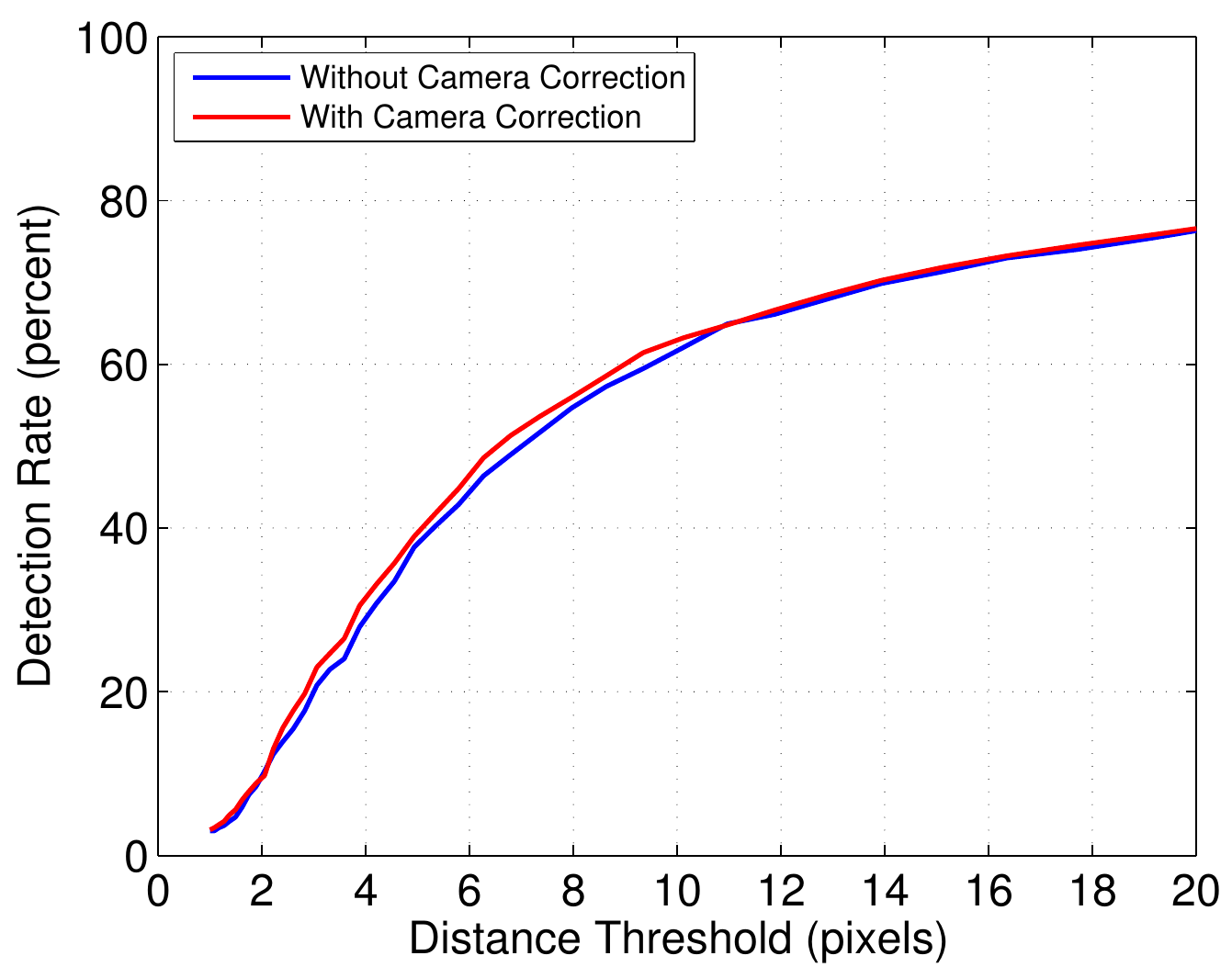}
  }
  \caption{Model performance for (a) varying motion feature frame offsets (b) with and without camera motion compensation}
  \label{fig:analysis}
\end{figure}

Fig~\ref{fig:analysis}(a) shows that the performance of our network is relatively agnostic to the frame separation ($\delta$) between the samples for which we calculate motion flow; the average precision between 0 and 20 pixel radii degrades 3.9\% from -10 pixels offset to -1 pixel offset.  A frame difference of 10 corresponds to approximately 0.42sec (at 24fps), and so we expect that large motions over this time period would result in complex non-linear trajectories in input space for which a single finite difference approximation of the pixel velocity would be inaccurate.  Accordingly, our results show that performance indeed degrades as a larger frame step is used.

Similarly, we were surprised that our camera motion compensation technique (described in section~\ref{sec:motionFeats}) does not help to the extent that we expected, as shown in Fig~\ref{fig:analysis}(b).  Likely this is because either LMN  removes a lot of constant background motion or the network is able to learn to ignore the remaining foreground-background parallax motion due to camera movement. 

\begin{figure}[h]
  \centering
  \subfloat[FLIC-motion: Elbow\label{fig:flic_elbow}]{
        \includegraphics[width=.48\textwidth]{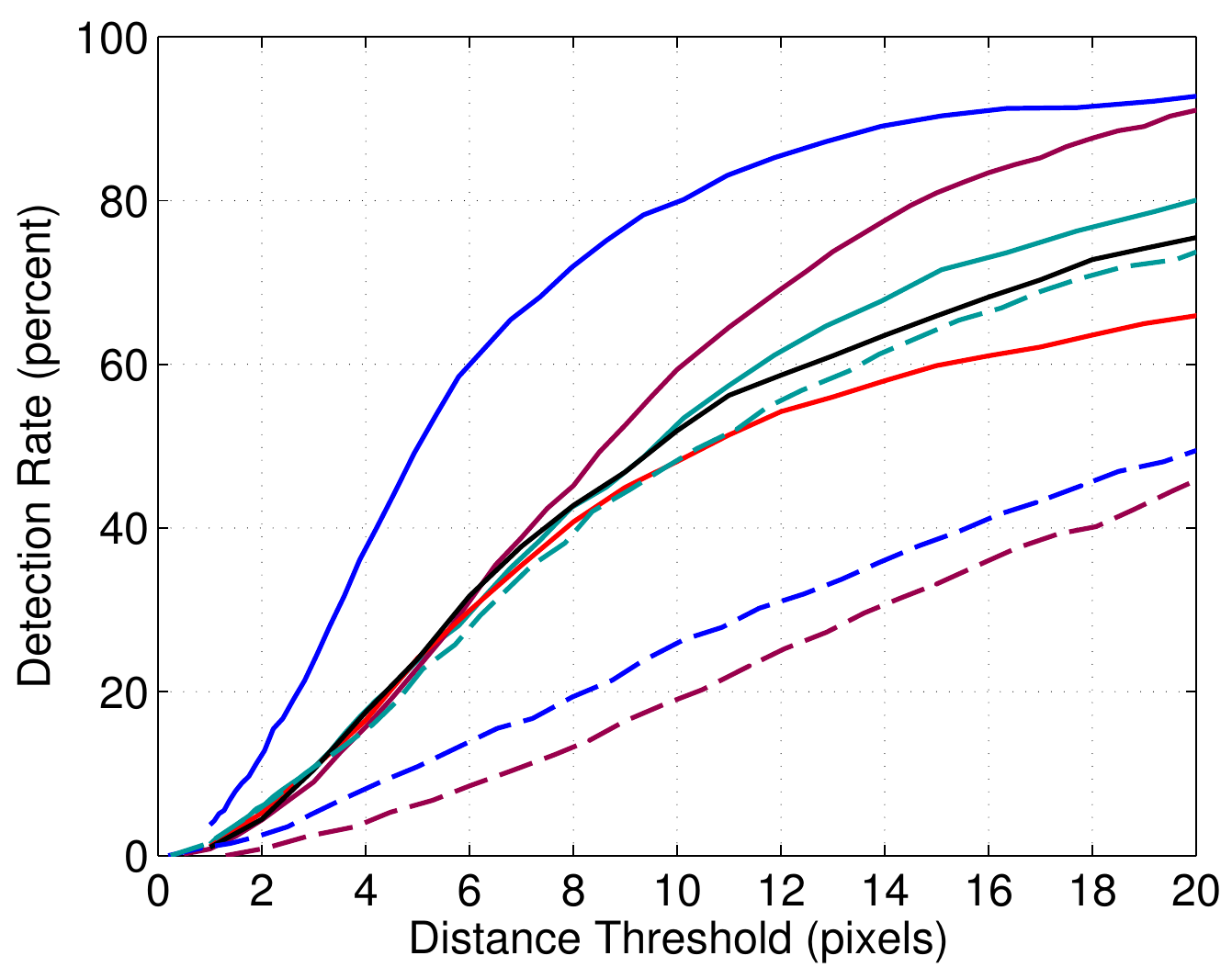}
   }
  \subfloat[FLIC-motion: Wrist\label{fig:flic_wrist}]{
        \includegraphics[width=.48\textwidth]{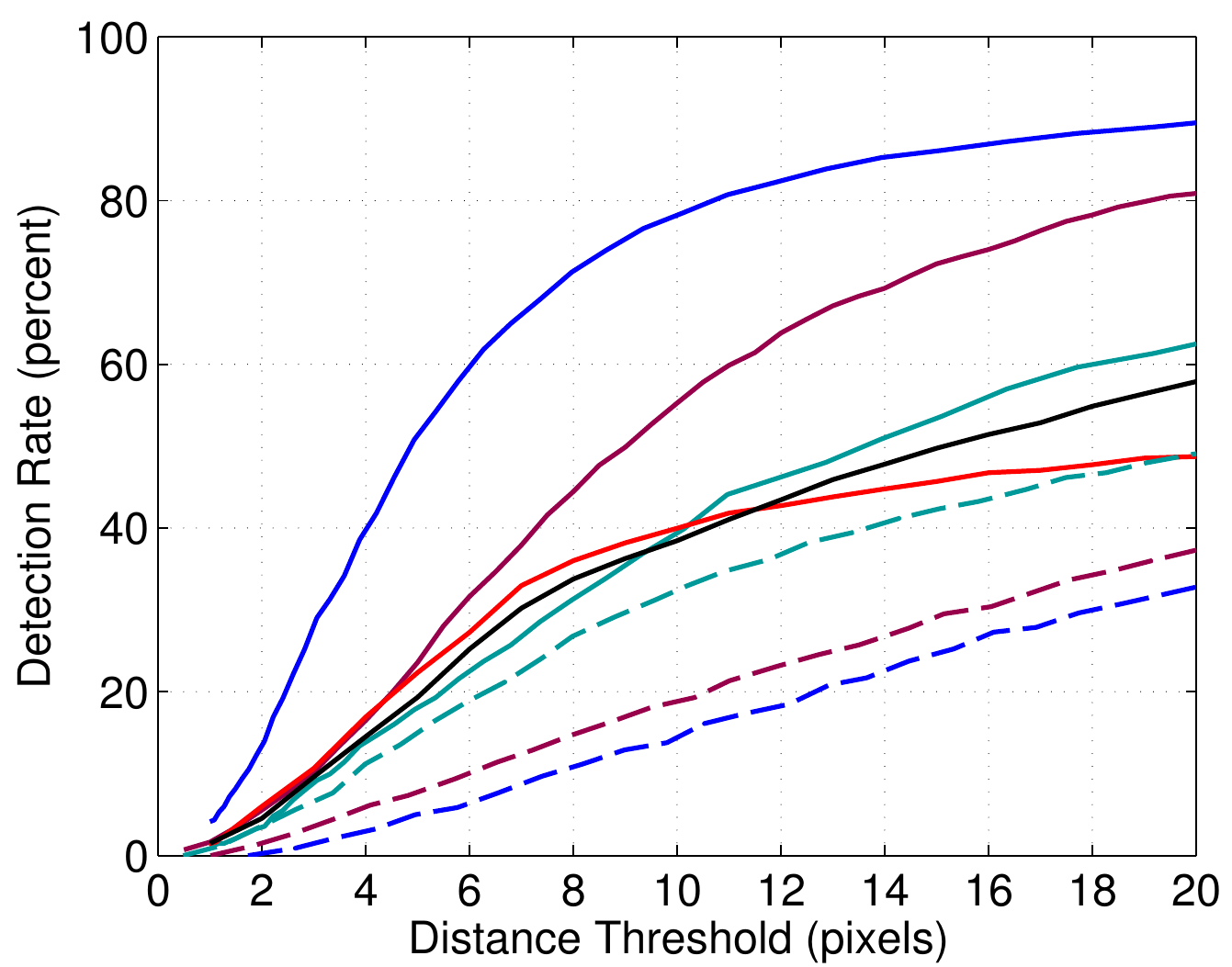}
   }\\
  \subfloat{
  \includegraphics[trim=0cm 0.6cm 1.5cm 0.6cm, clip=true, width=0.95\textwidth]{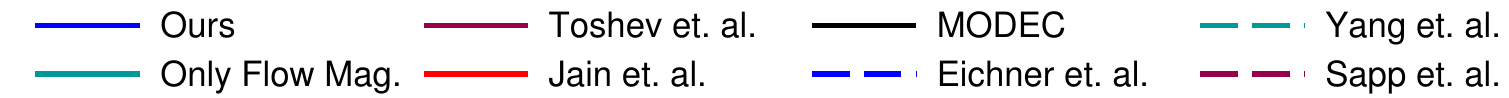}
   }
  \caption{Our model performance compared with our model using only flow magnitude features (no RGB image), Toshev et al.~\cite{deeppose}, Jain et al.~\cite{jainiclr2014},  MODEC~\cite{modec}, Eichner et al.~\cite{Eichner:2009:BAM}, Yang et al.~\cite{yang11cvpr} and Sapp et al.~\cite{sapp11eccv}.}
  \label{fig:flic_results}
\end{figure}
\subsection{Comparison with Other Techniques}
\label{sec:resultsComparison}

Fig~\ref{fig:flic_results}(a) and \ref{fig:flic_results}(b) compares the performance of our system with other state-of-the-art models on the FLIC dataset for the elbow and wrist joints respectively.  Our detector is able to significantly outperform all prior techniques on this challenging dataset. Note that using only motion features already outperforms~\cite{Eichner:2009:BAM, yang11cvpr, sapp11eccv}. Also note that using only motion features is less accurate than using a combination of motion features and RGB images, especially in the high accuracy region. This is because fine details such as eyes and noses are missing in motion features. Toshev et al.~\cite{deeppose} suffers from inaccuracy in the high-precision region, which we attribute to inefficient direct regression of pose vectors from images. MODEC~\cite{modec}, Eichner et al.~\cite{Eichner:2009:BAM} and Sapp et al.~\cite{sapp11eccv} build on hand crafted HoG features. They all suffer from the limitations of HoG (i.e. they all discard color information, etc). Jain et al.~\cite{jainiclr2014} do not use multi-scale information and evaluate their model in a sliding window fashion, whereas we use the `one-shot' approach. Finally, we believe that increasing the complexity of our simple spatial model will improve performance of our model, specifically for large radii.

\section{Conclusion}

We have shown that when incorporating both RGB and motion features in our deep ConvNet architecture, our network is able to outperform existing state-of-the-art techniques for the task of human body pose detection in video. We have also shown that using motion features alone can outperform some traditional algorithms~\cite{Eichner:2009:BAM, yang11cvpr, sapp11eccv}. Our findings suggest that even very simple temporal cues can greatly improve performance with a very minor increase in model complexity. As such, we suggest that future work should place more emphasis on the correct use of motion features.  We would also like to further explore higher level temporal features, potentially via learned spatiotemporal convolution stages and we hope that using a more expressive temporal-spatial model (using motion constraints) will help improve performance significantly.

\section{Acknowledgments}
The authors would like to thank Tyler Zhu for his help with the data-set creation. This research was funded in part by the Office of Naval Research ONR Award N000141210327.

\bibliographystyle{splncs}
\bibliography{mf_accv2014}

\end{document}